\documentclass[10pt,journal,compsoc]{IEEEtran}
\ifCLASSOPTIONcompsoc
  \usepackage[nocompress]{cite}
\else
  \usepackage{cite}
\fi
\usepackage{times}
\usepackage{threeparttable}
\usepackage{epsfig}
\usepackage{graphicx}
\usepackage{amsmath}
\usepackage{amssymb}
\usepackage{times}
\usepackage{epsfig}
\usepackage{graphicx}
\usepackage{amsmath}
\usepackage{amssymb}
\usepackage{multirow}
\usepackage{url}
\usepackage{caption}
\usepackage{subfigure}
\usepackage{float} 
\usepackage{algpseudocode,algorithm}
\usepackage{gensymb}
\usepackage{booktabs}
\usepackage{xcolor}
\usepackage{bm} 

\usepackage{amsmath,amsfonts,bm}

\def\eqref#1{equation~\ref{#1}}

\def\floor#1{\lfloor #1 \rfloor}
\def\1{\bm{1}}

\def\vtheta{{\bm{\theta}}}
\def\va{{\bm{a}}}

\def\ve{{\bm{e}}}

\def\vg{{\bm{g}}}

\def\vp{{\bm{p}}}

\def\vr{{\bm{r}}}

\def\vt{{\bm{t}}}

\def\vv{{\bm{v}}}

\def\vz{{\bm{z}}}

\def\mW{{\bm{W}}}

\DeclareMathAlphabet{\mathsfit}{\encodingdefault}{\sfdefault}{m}{sl}
\SetMathAlphabet{\mathsfit}{bold}{\encodingdefault}{\sfdefault}{bx}{n}

\def\sI{{\mathbb{I}}}

\def\sM{{\mathbb{M}}}

\def\sP{{\mathbb{P}}}

\def\sR{{\mathbb{R}}}
\def\sS{{\mathbb{S}}}

\DeclareMathOperator{\sign}{sign}

\usepackage{caption}
\usepackage{graphicx}
\usepackage{float}
\usepackage{algpseudocode,algorithm}
\usepackage{gensymb}
\usepackage{booktabs}
\usepackage{multirow}
\usepackage{stfloats}
\usepackage{graphicx,tabulary,amsmath,amssymb,booktabs,multirow,multicol,xspace,xcolor,bbding,comment,algorithm,listings,pifont,colortbl,makecell}
\graphicspath{{figure/}}
\definecolor{mygreen}{HTML}{39b54a}  
\definecolor{myred}{HTML}{ea4335}  
\definecolor{verylightgray}{gray}{0.9}

\newcommand{\better}[1]{\textcolor{mygreen}{#1}}
\newcommand{\worse}[1]{\textcolor{myred}{#1}}

\newcommand{\app}{\raise.17ex\hbox{$\scriptstyle\sim$}}

\newlength\savewidth

\makeatletter\renewcommand\paragraph{\@startsection{paragraph}{4}{\z@}
  {.5em \@plus1ex \@minus.2ex}{-.5em}{\normalfont\normalsize\bfseries}}\makeatother

\newcolumntype{x}[1]{>{\centering\arraybackslash}p{#1pt}}
\newcolumntype{a}[1]{>{\columncolor{verylightgray}\centering\arraybackslash}p{#1pt}}
\newcolumntype{y}[1]{>{\raggedright\arraybackslash}p{#1pt}}
\newcolumntype{z}[1]{>{\raggedleft\arraybackslash}p{#1pt}}

\ifCLASSINFOpdf
\else
\fi
\begin{document}
%
\title{{\rm AdPE}: Adversarial Positional Embeddings for Pretraining Vision Transformers via MAE+}
%
%
%
%

\author{Xiao Wang,~\IEEEmembership{Student~Member,~IEEE},
        Ying Wang, Ziwei Xuan,
        and Guo-Jun Qi,~\IEEEmembership{Fellow,~IEEE}
\IEEEcompsocitemizethanks{\IEEEcompsocthanksitem X. Wang is with Department of Computer Science, Purdue University,West Lafayette, 47906, USA.
\IEEEcompsocthanksitem Y. Wang is with Amazon AWS AI Labs, Seattle, WA 98121, USA. Email: yingwang0022@gmail.com.
\IEEEcompsocthanksitem Z. Xuan and G-J. Qi are with OPPO US Research Center, Seattle, WA, 98006, USA. Email: guojunq@gmail.com.
}
\thanks{Manuscript received March, 2023; revised xx, xx; accepted xx,xx.(Corresponding author: Guo-Jun Qi)}}

\IEEEtitleabstractindextext{%
\begin{abstract}
Unsupervised learning of vision transformers seeks to pretrain an encoder via pretext tasks without labels. Among them is the Masked Image Modeling (MIM) aligned with pretraining of language transformers by predicting masked patches as a pretext task. A criterion in unsupervised pretraining is the pretext task needs to be sufficiently hard to prevent the transformer encoder from learning trivial low-level features not generalizable well to downstream tasks.  For this purpose, we propose an Adversarial Positional Embedding (AdPE) approach -- It distorts the local visual structures by perturbing the position encodings so that the learned transformer cannot simply use the locally correlated patches to predict the missing ones. We hypothesize that it forces the transformer encoder to learn more discriminative features in a global context with stronger generalizability to downstream tasks. We will consider both absolute and relative positional encodings, where adversarial positions can be imposed both in the embedding mode and the coordinate mode. We will also present a new MAE+ baseline that brings the performance of the MIM pretraining to a new level with the AdPE. The experiments demonstrate that our approach can improve the fine-tuning accuracy of MAE by $0.8\%$ and $0.4\%$ over 1600 epochs of pretraining ViT-B and ViT-L on Imagenet1K. For the transfer learning task, it outperforms the MAE with the ViT-B backbone by $2.6\%$ in mIoU on ADE20K, and by $3.2\%$ in AP$^{bbox}$ and $1.6\%$ in AP$^{mask}$ on COCO, respectively. These results are obtained with the AdPE being a pure MIM approach that does not use any extra models or external datasets for pretraining. The code is available at \url{https://github.com/maple-research-lab/AdPE}.
\end{abstract}

}

\maketitle

\IEEEdisplaynontitleabstractindextext

%
\IEEEpeerreviewmaketitle

\IEEEraisesectionheading{\section{Introduction}\label{sec:introduction}}
\label{sec:intro}
Pretraining vision transformers \cite{dosovitskiy2020image} effectively has received many attentions due to the potential of unifying transformer architectures across modalities. Among them is the Masked Image Modeling (MIM) \cite{bao2021beit}\cite{he2022masked} that inherits the same idea in Bert pretraining of language transformers \cite{devlin2018bert}. These MIM approaches seek to predict the masked patches as a pretext task to pretrain vision transformers.

A critical principle in unsupervised pretraining of deep networks is the pretext task ought to be sufficiently hard to avoid trivial solutions only focusing on low-level features to bypass the task \cite{robinson2021can}. For this purpose, adversarial pretraining of the CNNs has demonstrated tremendous successes in the context of contrastive learning \cite{hu2021adco,kim2020adversarial,robinson2021can,wang2022caco}. For example, \cite{hu2021adco} learn to generate  hard negatives, forcing the network encoder to learn more discriminative features to distinguish hard negatives from their positive counterparts. By adopting harder pretext tasks \cite{robinson2021can}, the results showcase the adversarial pretraining is able to prevent the deep network from learning trivial low-level features with poor generalizability to downstream tasks.

Along this line of research, we aspire to develop an adversarial approach to effectively pretrain transformers in a MIM manner.  While the adversarial contrastive learning seeks to learn hard negatives \cite{hu2021adco,kalantidis2020hard,robinson2020contrastive}, no such adversaries exist in the MIM-based pretraining.  Thus, we first need to answer the question of what to choose as the adversary in the MIM-pretrained transformer. In this paper, we propose that perturbing the positional encodings is a natural choice to adversarially pretrain the vision transformer. It spatially distorts the local visual structures through perturbed positional encodings, and thus prevents the transformer from learning trivial features by exploiting the strong correlations between masked patches and their local unmasked peers.

In this way, we hypothesize the transformer is adversarially pretrained to focus on global contexts that are more useful for the downstream tasks to predict masked patches. We will consider adversarial perturbations applied in the positional embedding space additively, or on the image coordinates through a differentiable positional embedding/indexing function. Both absolute \cite{shaw2018self} and relative positional embeddings \cite{wu2021rethinking} will be considered for adversarial pretraining of transformers.

In addition, we will present a new MAE+ baseline by seamlessly fitting multi-crop tokenization to MAE \cite{he2022masked}.  While multi-crop augmentation has been successfully applied in the contrastive pretraining of both CNNs \cite{caron2020unsupervised} and transformers \cite{caron2021emerging}, we will demonstrate how this simple mechanism can be delicately designed with a lighter-weighted decoder having fewer input tokens than the existing MIM baseline \cite{he2022masked}. It allows to strike a better trade-off between the lighter decoder and more crops of tokenization to improve the MIM-pretraining.
We will demonstrate this results in a more efficient MAE+ baseline reaching superior performances alongside the proposed Adversarial Positional Embeddings (AdPE).

The remainder of this paper is organized as follows. We will review the related works in Section~\ref{sec:related}. Then we will revisit the key idea behind the MIM and present a new MAE+ baseline in Section~\ref{sec:multi}.  We will elaborate on the proposed Adversarial Positional Embeddings (AdPE) in Section~\ref{sec:adpe}. Experiment results will be reported in Section~\ref{sec:exp}, and we will conclude the paper in Section~\ref{sec:concl}.

\section{Related Works}\label{sec:related}
We will review the works that are closely related with the proposed method from three aspects -- MIM-based pretraining of transformers, absolute and relative positional encodings, and adversarial pretraining of deep networks.

\subsection{Masked Image Modeling and Pretraining Vision Transformers}
While it is natural to extend the contrastive learning approaches that achieve tremendous successes in pretraining the CNNs to pretrain the Vision Transformers (ViTs), Masked Image Modeling (MIM) \cite{bao2021beit}\cite{he2022masked} provides an alternative way inspired by the success in the NLP domain \cite{devlin2018bert}. This is expected to unify the transformer pretraining in computer vision, NLP and multmodality domains. The idea is simple -- it aims to unsupervisedly train a transformer encoder by masking out some tokens and reconstructing them at the decoder end. Such a pretext task aims to learn a useful transformer backbone as its encoder to learn representations that are useful
to predict the masked contents.

It is demonstrated that by masking out a large portion of input images (e.g., 75\% masked out \cite{he2022masked}), a hard MIM task is formulated that forces the encoder to learn useful clues in the long-range contexts to predict the missing patches. Masked tokens will be added either before \cite{bao2021beit} or after \cite{he2022masked} the encoder.  By only feeding the unmasked tokens through the encoder, the Masked Auto-Encoder (MAE) is able to reduce the computing costs for its encoder during the pretraining \cite{he2022masked}. However, it still needs to feed the masked tokens through the decoder to predict the missing patches.

\subsection{Positional Encodings}
The self-attention mechanism is unaware of position information by itself, and Positional Encodings (PEs) are thus required to represent the knowledge of where a token is in a sentence or in an image. Sinusoid embedding of positions was first proposed in the seminal paper \cite{vaswani2017attention}. It transforms a hard-coded position into a Fourier basis through sine/cosine functions at different frequencies. Such an Absolute PE (APE) is variant when the sequence is shifted.

Alternatively, Relative PEs (RPEs) \cite{shaw2018self,dai2019transformer,ramachandran2019stand} aim to encode the positions between tokens based on their relative relations, such as their relative distances and positions \cite{wu2021rethinking}. The resultant positional encodings are invariant when the sequence or the image is shifted. Various RPEs have been proposed. We will revisit them in Section~\ref{sec:rpe} before discussing how adversaries can be imposed on the RPEs.

\subsection{Adversarial Pretraining}
While adversarial {\em training} has been intensively studied through adversarial examples \cite{szegedy2013intriguing}\cite{goodfellow2014explaining}\cite{madry2017towards}, adversarial {\em pretraining}  of deep networks, especially CNNs, is receiving lots of attentions recently in the context of contrastive learning \cite{kim2020adversarial,robinson2021can}.  One of representative works in this line of research is to treat negative samples as adversarially learnable by maximizing instead of minimizing the InfoNCE loss \cite{hu2021adco}. Hard negatives are thus generated directly, where they are continuously being pushed towards their positive counterparts so that more discriminative features must be learned to distinguish between positives and negatives for a query. The idea was also extended to incorporate learnable positives in a cooperative-adversarial fashion together with learnable negatives \cite{wang2022caco}. All these methods focus on the adversarial pretraining for contrastive learning.

In this paper, we seek to adversarially pretrain vision transformers in the MIM fashion. In contrast to obtaining hard negatives in \cite{hu2021adco,kalantidis2020hard,robinson2020contrastive}, we will impose adversarial perturbations on positional embeddings to distort the local visual structures spatially so that the pretrained transformers are forced to explore high-level features in long-range contexts to predict missing patches, instead of simply cheating on the strongly correlated patches in local neighborhood.


\section{Masked Image Modeling and A New MAE+ Baseline} \label{sec:multi}
In this section, first we will briefly revisit the Masked Image Modeling (MIM)-based approaches for pretraining vision transformers. Then we will present a new MIM-pretraining baseline, {\bf MAE+} based on multi-crop tokenization that seamlessly fits the state-of-the-art MAE \cite{he2022masked}, boosting performances with less computing cost. The MAE+ will be used as the new baseline to showcase the greater potential of the MIM-pretraining.

\subsection{Masked Image Modeling and Masked Auto-Encoders}

The Masked Image Modeling (MIM)-based methods pretrain vision transformers through an encoder-decoder architecture.  Given an image, it is flattened to a sequence of tokens corresponding to a group of non-overlapping patches. A large portion of patch tokens in the input sequence will be masked out, being replaced with a mask token before feeding into the encoder \cite{bao2021beit}. Alternatively, only unmasked patch tokens will be input to the encoder \cite{he2022masked}, and a shared learnable mask token will be used to represent each masked patch before feeding all tokens into the decoder to predict the missing patches.

The decoder can be composed of several layers of transformers \cite{he2022masked}, or simply consist of a simple fully-connected layer \cite{xie2022simmim}.  The encoder is a backbone transformer that we wish to pretrain and use later in the downstream tasks. A mean-squared reconstruction error over masked patches is minimized to pre-train both the encoder and the decoder end-to-end through back-propagation in the Masked Auto-Encoder (MAE) architecture \cite{he2022masked}.

\subsection{MAE+: A New Baseline for MIM-Pretraining of Vision Transformers}\label{sec:mae+}

Typically, in a MIM-based approach \cite{bao2021beit,he2022masked}, an input image of resolution $224\times 224$ is divided into $14\times 14$ tokens, each of which corresponds to a non-overlapping patch of $16\times16$.  In the MAE \cite{he2022masked}, $75\%$ tokens are masked out, leaving only $49$ unmasked tokens that will be fed through an ViT encoder. After that, a group of learnable masked tokens will be concatenated  with the unmasked tokens output from the encoder, forming a total of $196$ tokens which will then go through several layers of transformer decoders to reconstruct the patches of masked tokens.

Such a MAE architecture has some drawbacks preventing it from releasing its full potentials. First, although only a much smaller number of $49$ unmasked tokens are fed into the encoder, the decoder still has a full number of $196$ tokens as input. The computational complexity squared in the number of the decoder tokens still incurs heavy costs to pretrain a ViT network. Second, Although multi-crop augmentation has been studied in both CNNs \cite{caron2020unsupervised} and vision transformers \cite{caron2021emerging} for contrastive learning, a delicate mechanism tailored to the masked image modeling has yet to be developed considering both efficiency and accuracy.


To this end, we propose a new MIM baseline named MAE+, and show that it can seamlessly fit the MIM-pretraining of transformers.
Particularly, we randomly crop a full-sized image by half to a small scale of $112\times112$, but still tokenize it with non-overlapping patches of $16\times16$. This will result in exactly $49 (=7\times7)$ tokens, the same number of unmasked tokens fed into the MAE encoder. Then we randomly mask out $75\%$ tokens. Unlike the MAE, we allow all these $49$ tokens to feed through the encoder, no matter if they are masked or not, which does not incur more computational burden than the MAE encoder.  Now, the decoder will only take these $49$ tokens as input. In contrast, the MAE decoder has a total of $196$ masked and unmasked tokens as its input on a full-sized image, whose computing complexity is up to $16$ times that of a cropped one.

\begin{figure}
\begin{center}
\includegraphics[width=0.48\textwidth]{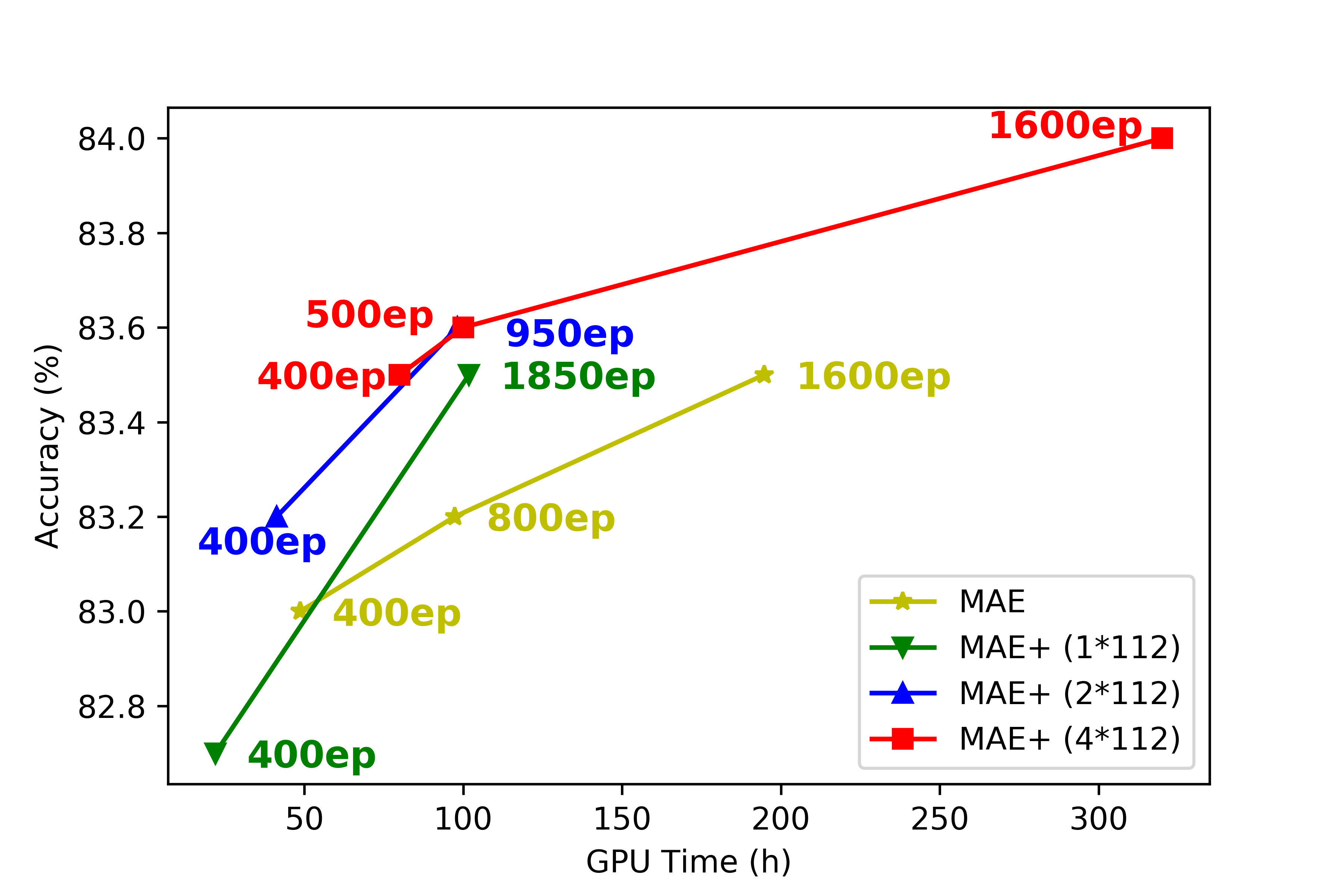}
\end{center}
\caption{The figure illustrates the top-1 accuracy on Imagenet1K achieved by MAE and MAE+ with the ViT-B backbone. It compares the accuracy against the GPU hours (on a single GPU server with eight V100 Nvidia cards) used for network pretraining. The $k*112$ in the parenthesis denotes $k$ of $112*112$ sub-images are cropped to tokenize an input image. Note that the pretrained backbone is fine-tuned on the original $224*224$ images without multi-crop tokenization following the evaluation protocol \cite{he2022masked}. }\label{fig:MAEPlus}

\end{figure}

The saved computational cost in the decoder allows us to have multiple crops without increasing the computing overhead. One can even adopt a simple fully-connected layer as the decoder as in the SimMIM \cite{xie2022simmim}. More crops of tokenization can yield better performances, and a delicate trade-off can be reached between the lower complexity of the decoder and more tokenized crops per image to pretrain a vision transformer to its great potential.  Once pretrained, the transformer encoder is used conventionally on full-sized images as a backbone in downstream tasks.

Figure~\ref{fig:MAEPlus} compares the top-1 accuracy achieved by MAE and MAE+ with the ViT-B backbone by pretraining on Imagenet1k dataset.   The same encoder and decoder transformers are adapted except the MAE+ uses the multi-crop tokenization. The same evaluation protocol as in MAE \cite{he2022masked} is adopted by fine-tuning the pretrained encoder with Imagenet1K labels on the original $224*224$ images {\em without} multi-crop tokenization. The results show that MAE+ can remarkably improve the accuracy while reducing the computing cost.  For example, with half of GPU hours, even a single crop MAE+ (1*112) can achieve the same top-1 accuracy as the MAE pretrained for $1600$ epochs. With more crops $k$, more pretraining cost can be saved with even higher accuracy. The details about the experiment setting on the MAE+ are discussed in Section.~\ref{sec:implement}. In the following, we will adopt MAE+ with four crops ($4*112$) as the baseline model for the proposed approach.

\section{Adversarial Positional Embeddings}\label{sec:adpe}
\label{sec:ad}

In this section, we will present the proposed Adversarial Positional Embeddings (AdPE). On a high level, it seeks to make it harder to reconstruct the masked tokens. In particular, the AdPE will distort the local spatial structures, which will prevent the MIM from merely leveraging the local correlations between tokens to reconstruct those masked ones. In this way, the learned representation can capture more useful high-level features in global contexts.

Adversarial perturbations will be added onto the positional embeddings in two ways: in the embedding space (i.e., embedding mode) and in the spatial coordinates (i.e., coordinate mode). We will consider both absolute positional embeddings and relative positional embeddings for the AdPE.

\subsection{Adversarial Absolute Positional Embeddings}

Consider a token representation $\vt_i\in \sR^d$ at an image location $(x_i,y_i)$. Before it is fed through a transformer, a positional encoding $\vp_{x_i,y_i}$ of its coordinates is added onto $\vt_i$ to give rise to a resultant position-aware token. 

We seek to distort the local structures represented by the positional embeddings, so that the MIM cannot merely explore the patch-level correlations in a local affinity to predict the masked tokens.  For this, there are two different ways to add adversarial perturbations to the positional information.

\subsubsection{Embedding Mode Adversaries}

The most straight way is to directly add an adversarial perturbation $\va\in \sR^d$ onto the positional embedding $\vp_{x_i,y_i}$. This results in a perturbed positional embedding $\vp_{x_i,y_i} + \va$ being fed into a transformer layer. Leaving all the trainable network weights in $\vtheta$, the MIM objective becomes
$$
\min_{\vtheta}\max_{\|\va\|_q\leq \epsilon} \mathcal L(\{\vp_{x_i,y_i} + \va|i\in\sM\};\vtheta)
$$
where $\mathcal L$ is the MAE loss that minimizes the mean squared reconstruction errors over masked tokens, and $\|\va\|_q \leq \epsilon$ is the $\ell_q$-norm constraint on the magnitude of the perturbation.
The adversarial perturbations are only added to the set $\sM$ of masked tokens, and better performances have been observed by sharing adversaries $\va$ over masked tokens. To solve this constrained minimax problem, we will adopt the following two strategies throughout the paper.

{\noindent \bf Parallel mode updates.} We adopt the stochastic gradient to update $\vtheta$ and $\va$. However, unlike the other adversarial approach \cite{szegedy2013intriguing} that sequentially update two adversarial players, these two parts are {\em simultaneously} updated via negative gradient (i.e., minimizing the loss over $\vtheta$) and  positive gradient (i.e., maximizing the loss over $\va$). In other words, for each iteration, we only feed a sample forward and backward {\em once} through the network to update $\vtheta$ and $\va$ in parallel, without alternating between them \footnote{In sequential mode, each time after updating and fixing one part of parameters, one needs to feed forward and backward an image again before the other part being updated and fixed.}.  We do not find any significant difference between sequential and parallel updates in performance but the latter can save up to half of the computing time per iteration. 

{\noindent \bf Projected gradient descent.} The $\ell_q$-norm constraint can be enforced by the Projected Gradient Descent (PGD) iteratively \footnote{Alternatively, linearizing the loss $\mathcal L$ can lead to a direct update to $\va$ \cite{goodfellow2014explaining}\cite{kurakin2016adversarial}. For the $\ell_2$ norm, we have $\va = \frac{\epsilon\vg}{\|\vg\|_2}$, and for the $\ell_\infty$ norm, $\va = \epsilon{\rm sign}(\vg)$, where $\vg = \nabla_\va \mathcal L(\{\vp_{x_i,y_i} + \va\}_i;\vtheta)|_{\va=\mathbf 0}$ is the loss gradient at zero perturbation. However, the iterative update via PGD can find more sophisticated adversarial perturbations than the direct approach \cite{tramer2017space}.}. Once $\va$ is updated each time, it is immediately projected onto the $\ell_q$-ball with a radius of $\epsilon$ to meet the constraint, that is
$$
\va \leftarrow \Pi_{\sS} {(\va+\alpha\nabla_{\va}\mathcal L(\{\vp_{x_i,y_i} + \va|i\in\sM\};\vtheta))}
$$
with a learning rate $\alpha$, and the constraint set $\sS = \{\va|\|\va\|_q\leq \epsilon\}$.
For the $\ell_2$ norm, the projection clips the resultant perturbation to a maximum length of $\epsilon$, i.e., $\Pi_{\sS} {(\vv)} = \frac{\min(\epsilon, \|\vv\|_2)\cdot\vv}{\|\vv\|_2}$. For the $\ell_\infty$ norm, the projection clips the resultant perturbation element-wise to a maximum absolute value of $\epsilon$, i.e., $[\Pi_{\sS} {(\vv)}]_l= \min(|v_l|,\epsilon) \cdot {\rm sign} (v_l)$.

We will adopt both strategies to update the adversarial perturbation unless stated otherwise. In experiments, the computing overhead for the PGD-based AdPE with the parallel mode updates is less than $1.5\%$ compared to the MAE+ baseline.

\subsubsection{Coordinate Mode Adversaries}

Adding adversaries in the positional embedding space is implicit in how the underly spatial structure is perturbed.  A more direct way is to add the adversaries $\boldsymbol\delta=(\delta_x,\delta_y)$ to the underlying coordinates, resulting in disturbed coordinates $(x_i+\delta_x, y_i+\delta_y)$ and the corresponding positional embedding $\vp_{x_i+\delta_x, y_i+\delta_y}$.

In this case, the objective of the MIM task becomes
$$
\min_{\vtheta}\max_{\|\boldsymbol\delta\|_q \leq \epsilon} \mathcal L(\{\vp_{x_i+\delta_x, y_i+\delta_y}|i\in\sM\};\vtheta)
$$
such that the adversaries are maximized to distort the spatial coordinates.  Usually the absolute positional embedding is a sinusoid function \cite{vaswani2017attention},
$$
\vp_{x_i+\delta_x, y_i+\delta_y} = \begin{bmatrix}\sin((x_i+\delta_x)/10000^{2j/d_{\rm model}}) \\\cos((x_i+\delta_x)/10000^{2j/d_{\rm model}}) \end{bmatrix}_j 
$$
$$
\oplus \begin{bmatrix}\sin((y_i+\delta_y)/10000^{2j/d_{\rm model}}) \\\cos((y_i+\delta_y)/10000^{2j/d_{\rm model}}) \end{bmatrix}_j
$$
which is differential in $\boldsymbol\delta$ so it can be learned via back-propagation. Here, $d_{\rm model}$ is the dimension compatible with the model, and $\oplus$ denotes the vector concatenation.

\subsection{Adversarial Relative Positional Embeddings}
Like in absolute positional embeddings, the relative positional embeddings also have two adversarial modes on embeddings and coordinates, respectively.

\subsubsection{Relative Positional Encodings}\label{sec:rpe}
Let us first revisit the relative positional encodings \cite{wu2021rethinking}.  Consider a pair of token representations $\vt_i$ and $\vt_j$. We have a scaled correlation matrix $\ve$ whose entries $e_{ij}$ is computed as
$$
e_{ij}=\dfrac{(\vt_i \mW^Q \cdot \vt_j \mW^K)+b_{ij}}{\sqrt{d_z}}
$$
where the bias is either $b_{ij}=r_{ij}$ or $b_{ij}=\vt_i \mW^Q \cdot \vr_{ij}$ with $\vr_{ij}\in\sR^{d_z}$.

Here $\vr_{ij}$ is the relative positional encoding for a token pair $\vt_i$ and $\vt_j$, which is a learnable vector. A softmax function is applied to transform $e_{ij}$ into an attention matrix $\boldsymbol\alpha$, and the output token can be written as
$$
\vz_i = \sum_{j=1}^n \alpha_{ij} (\vt_j \mW^V + \vr^V_{ij})
$$
with $\vr^V_{ij}\in \sR^{d_z}$ is a value embedding of relative positions between the two tokens. For simplicity, we can use a unified representation $\vr_{ij}$ to denote the positional embeddings in different cases.

An index function ${\rm Ind}:\sI\times\sI\rightarrow\sI$ is defined to map a pair of tokens $(i,j)$ to an integer index ${\rm Ind}(i,j)$ such that a learnable positional embedding $\vp_{{\rm Ind}(i,j)}$ can be retrieved from a dictionary. For an image, its x-coordinate and y-coordinate will be mapped separately to two indices denoted by ${\rm Ind}_x(i,j)$ and ${\rm Ind}_y(i,j)$. Then the relative positional encoding $\vr_{ij}$ of the two tokens is given by 
$$\vr_{ij}\triangleq\vp_{[{\rm Ind_x}(i,j),{\rm Ind_y}(i,j)]}$$ 
with such a 2D index $[{\rm Ind_x}(i,j),{\rm Ind_y}(i,j)]$ in square brackets to a dictionary of relative positional embeddings $\sP=\{\vp_{[k,l]}|k,l=-\beta,\cdots,+\beta\}$ (see details below). 

\subsubsection{Embedding Mode Adversaries}
It is straight to add adversarial perturbation to the relative positional encodings directly,
$$
\vp_{[{\rm Ind_x}(i,j),{\rm Ind_y}(i,j)]}+\va,
$$
where $\va$ is such an additive adversarial perturbation in the embedding mode as in the aforementioned APE case.  Now the MIM task becomes
$$
\min_{\boldsymbol\theta,\vp}\max_{\|\va\|_q \leq \epsilon}\mathcal L(\{\vp_{[{\rm Ind_x}(i,j),{\rm Ind_y}(i,j)]}+\va|i\in\sM\};\vtheta)
$$
where the constrained adversarial perturbation $\va$ is applied so long as the first token is masked in a pairwise relation no matter if the second one is masked or not. While the positional embeddings $\vp$'s are learned by minimizing the MAE loss as the other weights $\vtheta$, the perturbation is learned in an adversarial manner by maximizing the loss. Relative positional encodings are applied to various layers of transformers. For each layer of transformer, a distinct adversarial perturbation is applied and learned. This allows us to perturb positional structures between tokens differently to model the transformer representation on various scales.
\begin{table*}[!th]
\begin{center}
\caption{Comparison of the AdPE with MAE+. The results show that the AdPE makes consistent improvement (in $\Delta$) over the MAE+ baseline.}
\label{tab:adv_mae+}

\begin{tabular}{lcccccc|lcccccc}
\toprule[1.5pt]
&&& \multicolumn{2}{c}{400ep}&\multicolumn{2}{c}{1600ep}&&&&\multicolumn{2}{c}{400ep}&\multicolumn{2}{c}{1600ep}\\
 \cline{4-7}\cline{11-14}
Method & PE & Mode & FT&$\Delta$&ft&$\Delta$&Method & PE & Mode & FT&$\Delta$&ft &$\Delta$\\
\midrule[1.5pt]
MAE+ & APE& - &83.51&-&83.94&- &MAE+ & RPE& - &83.78&-&84.17 &-\\
AdPE & APE &Embed& 83.67&\better{+0.16}&84.08&\better{+0.14}&AdPE & RPE &Embed&83.98&\better{+0.20}&84.25&\better{+0.08}\\
AdPE & APE &Coord &83.68&\better{+0.17}&84.16&\better{+0.22}& AdPE & RPE &Coord &84.01&\better{+0.23}&84.36&\better{+0.19}\\
\bottomrule[1.5pt]
\end{tabular}
\end{center}

\end{table*}
\subsubsection{Coordinate Mode Adversaries}

\begin{figure}
  \begin{center}
    \includegraphics[width=0.4\textwidth]{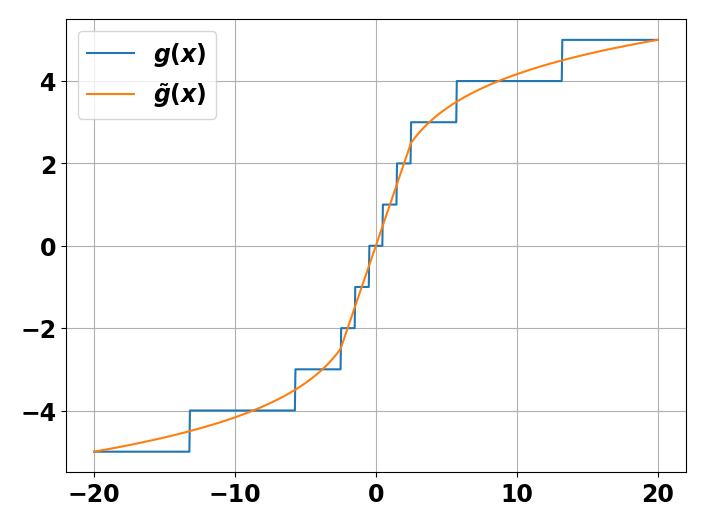}
  \end{center}
  \caption{Illustration of the integer-valued index function $g$ and its relaxed form $\tilde g$.}\label{fig:gfunc}
\end{figure}

%

The adversarial perturbations can also be added to the coordinates $(x_i,y_i)$ and $(x_j,y_j)$ directly.
 For example, a piece-wise function $g:\sR\rightarrow \sI$ in \cite{wu2021rethinking} has been introduced to define the index function such that ${\rm Ind_x}(i,j)=g(x_i-x_j)$ and ${\rm Ind_y}(i,j)=g(y_i-y_j)$, which is able to distribute various ranges of attention by the relative distance between two tokens $i$ and $j$ as illustrated in Figure~\ref{fig:gfunc}. Then the adversarial perturbations $\delta_x$ and $\delta_y$ on the relative coordinates in these two index functions give rise to an adversarial positional embedding $\vp_{[g(x_i-x_j+\delta_x),g(y_i-y_j+\delta_y)]}$.

However, the gradient of the adversarial positional embedding $
\vp_{[g(x_i-x_j+\delta_x),g(y_i-y_j+\delta_y)]}$ over the perturbation $\boldsymbol\delta$ is ill-defined for back-propagation for the step-wise $g$. A small perturbation on the image coordinates either incurs an abrupt change in indexing $g$ to another positional embedding, resulting in an infinitely large gradient; or makes no change at all, leading to a vanishing gradient.

To mitigate this problem, we relax the integer index $g$ to a real-valued function $\widetilde g$. For example, by removing the round operation, the piece-wise linear function is relaxed to
$$
\widetilde g(x)=\begin{cases}x & |x| \leq \alpha\\ \sign(x)\cdot \min(\beta,\alpha+\dfrac{\ln(|x|/\alpha)}{\ln(\gamma/\alpha)}(\beta-\alpha)) & |x| > \alpha\end{cases}
$$
Figure~\ref{fig:gfunc} compares the integer index $g$ with its relaxed form $\widetilde g$.

\begin{figure}
  \begin{center}
    \includegraphics[width=0.3\textwidth]{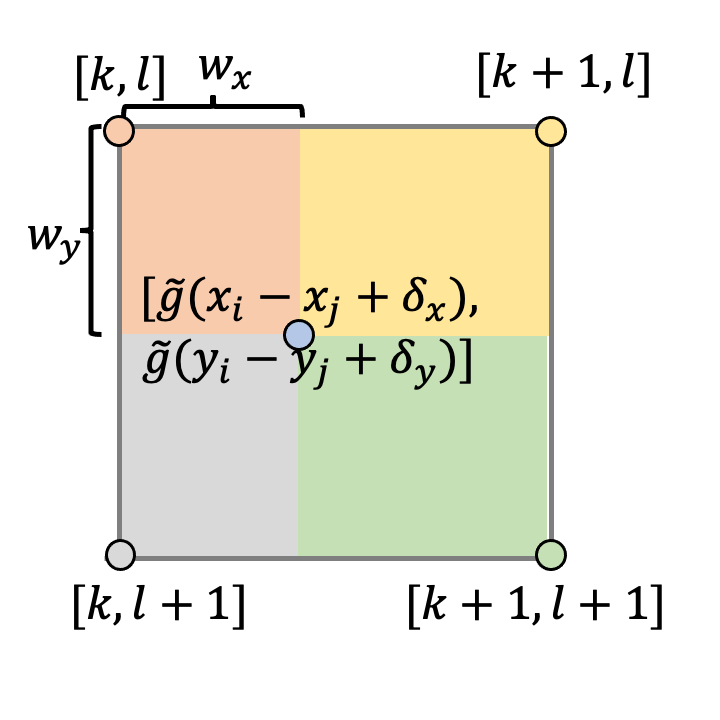}
  \end{center}
  \caption{The bilinear interpolation of four nearest positional embeddings via $\tilde g$.}\label{fig:bilinear}
\end{figure}

\begin{table*}[!th]
\caption{Top-1 fine-tuning (FT) accuracy over 400 epochs of pretraining ViT-B with different positional encodings (PE), adversarial modes (Mode), constraint types (Constraint), and the cutoff $\epsilon$ of constraint strength  (Cutoff). It also gives the relative improvement over the baseline MAE model. Here, APE and RPE stand for absolute and relative positional embeddings, respectively.}
\label{tab:ablation}
\begin{center}

\begin{tabular}{lccccc|lccccc}
\toprule[1pt]
PE & Mode& Constraint& Cutoff& FT& $\Delta_{MAE}$ &PE & Type& Constraint& Cutoff& ft& $\Delta_{MAE}$\\
\midrule[1pt]
APE &Embed& $\mathcal{\ell}_{2}$ & 1& 83.55 & \better{+0.60} & RPE&Embed&$\mathcal{\ell}_{2}$ & 3 &83.85&\better{+0.90}\\
APE & Embed&$\mathcal{\ell}_{2}$  & 3& 83.60 & \better{+0.65}& RPE&Embed&$\mathcal{\ell}_{2}$ & 5 &83.85&\better{+0.90}\\
APE &Embed&  $\mathcal{\ell}_{2}$  & 5& 83.63 &\better{+0.68}& RPE&Embed&$\mathcal{\ell}_{2}$ & 7 &83.82&\better{+0.87}\\
APE & Embed& $\mathcal{\ell}_{\infty}$  & 1& 83.53&\better{+0.58}& RPE&Embed&$\mathcal{\ell}_{\infty}$ & 10 &83.92&\better{+0.97}\\
APE & Embed& $\mathcal{\ell}_{\infty}$ & 3& \textbf{83.67}&\better{+0.72}& RPE&Embed&$\mathcal{\ell}_{\infty}$ & 15 &83.97&\better{+1.02}\\
APE & Embed& $\mathcal{\ell}_{\infty}$& 5& 83.62&\better{+0.67}& RPE&Embed&$\mathcal{\ell}_{\infty}$ & 20 &\textbf{83.98}&\better{+1.03}\\
\midrule
APE & Coord& $\mathcal{\ell}_{2}$& 1& 83.66 & \better{+0.71} & RPE&Coord&$\mathcal{\ell}_{2}$ & 3 & 81.13&\worse{-1.82}\\
APE & Coord&$\mathcal{\ell}_{2}$& 3& 83.54& \better{+0.59}& RPE&Coord&$\mathcal{\ell}_{2}$ & 5 & 81.18&\worse{-1.77}\\
APE & Coord& $\mathcal{\ell}_{2}$& 5& 83.59& \better{+0.64}& RPE&Coord&$\mathcal{\ell}_{2}$ & 7 & 81.21&\worse{-1.74}\\
APE & Coord& $\mathcal{\ell}_{\infty}$ & 1& 83.50& \better{+0.55}& RPE&Coord&$\mathcal{\ell}_{\infty}$ & 10 & 83.73&\better{+0.78}\\
APE & Coord& $\mathcal{\ell}_{\infty}$ & 3& \textbf{83.68}& \better{+0.73}& RPE&Coord&$\mathcal{\ell}_{\infty}$ & 15 &83.86&\better{+0.91}\\
APE & Coord& $\mathcal{\ell}_{\infty}$ & 5& 83.61&\better{+0.66}& RPE&Coord&$\mathcal{\ell}_{\infty}$ & 20 &\textbf{84.01}&\better{+1.06}\\
\bottomrule[1pt]
\end{tabular}
\end{center}
\end{table*}

To have differentiable positional embeddings over $\boldsymbol\delta$, we view $\left[\widetilde g(x_i-x_j+\delta_x),\widetilde g(y_i-y_j+\delta_y)\right]$ as a continuous 2D index, and apply bilinear interpolation to continuously retrieve the positional embedding from $\sP=\{\vp_{[k,l]}|k,l=-\beta,\cdots,+\beta\}$, the dictionary of all positional embeddings at integer 2D indices (see Figure~\ref{fig:bilinear}). Suppose the top-left corner nearest to the 2D coordinate has an integer index $[k,l]$, where $k=\floor{\widetilde g(x_i-x_j+\delta_x)}$ and $l=\floor{\widetilde g(y_i-y_j+\delta_y)}$ with the floor function $\floor{\cdot}$. Then it yields the interpolated adversarial positional embedding below under the perturbation $\boldsymbol\delta=(\delta_x,\delta_y)$
\begin{equation}
\begin{aligned}
&\vp_{[\widetilde g(x_i-x_j+\delta_x),\widetilde g(y_i-y_j+\delta_y)]} = (1-w_x)(1-w_y) \vp_{[k,l]} \\
&+ w_x(1-w_y) \vp_{[k+1,l]} + (1-w_x)w_y \vp_{[k,l+1]} \\
&+ w_x w_y \vp_{[k+1,l+1]}\nonumber
\end{aligned}
\end{equation}
with the interpolation weights
$$
w_x = \widetilde g(x_i-x_j+\delta_x)-k, w_y = \widetilde g(y_i-y_j+\delta_y)-l.
$$

The index function $\widetilde g$ and the coordinate perturbation $(\delta_x,\delta_y)$ appear in the bilinear weights, where the back-propagated errors can go through $\widetilde g$ via these weights to update the perturbation.


Finally, the MIM objective with adversarial coordinates on relative positional embeddings becomes
$$
\min_{\boldsymbol\theta,\vp}\max_{\|\boldsymbol\delta\|_q\leq\epsilon}\mathcal L(\{\vp_{[\widetilde g(x_i-x_j+\delta_x),\widetilde g(y_i-y_j+\delta_y)]}|i\in\sM\};\vtheta)
$$
with the constrained coordinate perturbation $\boldsymbol\delta$.

\section{Experiments}\label{sec:exp}
In this section, we present the experiment results for the proposed AdPE method.  We will show the results on Imagenet1K dataset with the pretrained model, the transfer learning results on other datasets, and ablation studies for AdPE, as well as visualize the attention maps learned by the AdPE model.

\subsection{Implementation Details}\label{sec:implement}

In Section~\ref{sec:mae+}, we propose a new MIM baseline MAE+ by allowing multi-crop tokenization. Specifically, on Imagenet1K dataset, following random cropping and resizing \cite{}, a $224*224$ input image is cropped into multiple $112*112$ smaller images. The $75\%$ of $7*7$ tokens resulting from each $112*112$ image with a grid of $16*16$ non-overlapping patches are randomly masked. Unlike the MAE, both masked and unmasked tokens will be fed through the MAE+ encoder.

MAE+ adopts the same encoder-decoder architecture as MAE~\cite{he2022masked}. The reconstruction loss over tokens is minimized to pretrain the network over various epochs. All experiments are run on a GPU server equipped with eight Nvidia V100 cards.

More specifically, in the pretraining stage, we follow the MAE baseline \cite{he2022masked}. The network is trained with the same linear learning rate scaling rule is applied - the base learning rate of $1.5e-4$ is adapted with a batch size of $4096$, yielding a learning rate of $lr=base_{lr}\times batchsize/256 = 2.4e-3$.  AdamW optimizer is used with its momentum parameters of $\beta_1=0.9$ and $\beta_2=0.95$. The cosine decay is also adopted for scheduling learning rate with $40$ warmup epochs and a weight decay of $0.05$.

In the fine-tuning stage,we also use the common practice \cite{he2022masked}\cite{bao2021beit} to supervise the end-to-end re-training of ViT-B. The fine-tuning evaluation is made by re-training the whole network with Imagenet labels where a linear classification layer is added upon the average-pooled features from the pretrained ViT backbone. A learning rate of $4e-3$ is used with a batch size of $1024$ and the cosine decay. AdamW is still used with the optimizer momentum set to $\beta_1=0.9$ and $\beta_2=0.999$. A layer-wise learning rate decay of $0.75$ is adopted. A total of $100$ epochs is adopted for the fine-tuning with $5$ warmup epochs. A drop path of $0.1$ is adopted and RandAug (9, 0.5) is used for data augmentation along with mixup, cutmix and label smoothing.

\begin{table*}[t]
\caption{Comparison of our model with other methods on ViT-B and ViT-L. We evaluate them with the top-1 fine-tuning accuracy on ImageNet.}
\label{tab:advall}

\begin{center}

\begin{tabular}{lccccccc}
\toprule[1.5pt]
Method & Type &Extra Model &Epochs&ViT-B& ViT-L \\
\midrule[1.5pt]
supervised~\cite{he2022masked}& Supervised &-&300&82.3&82.6\\
MoCo-v3~\cite{chen2021empirical}&Contrastive &momentum ViT & 300&83.2&84.1\\
DINO~\cite{caron2021emerging}&Contrastive &momentum ViT&300&82.8&-\\
iBOT~\cite{zhou2021ibot}&Contrastive+MIM & momentum ViT&1600& 84.0&84.8\\
BEiT~\cite{bao2021beit}&MIM &DALLE+dVAE & 800& 83.2&85.2\\
data2vec~\cite{baevski2022data2vec}&Contrastive& momentum ViT&800&84.2&86.2\\
CAE~\cite{chen2022context} & MIM &DALLE tokenizer&1600&83.9&86.3\\

\midrule
SimMIM~\cite{xie2022simmim} &MIM&-&800&83.8&85.4\\
MaskFeat~\cite{wei2022masked} &MIM&-&1600&84.0&85.7 \\
MAE~\cite{he2022masked}&MIM &-&1600& 83.6&85.9 \\

\midrule
MAE+ (ours) & MIM & - &1600& 83.9 & 86.0\\
AdPE (ours) &MIM  & - &1600 & {\bf 84.4} &{\bf 86.3}\\
\bottomrule[1.5pt]
\end{tabular}
\end{center}

\end{table*}
\subsection{Imagenet1K Results}

We adopt the MAE+ baseline presented in Section~\ref{sec:mae+} as our baseline model for MIM pretraining.  For the fair comparison, the same set of hyperparameters used in Section.~\ref{sec:implement} are adopted. There are four combinations of design choices as discussed in Section~\ref{sec:adpe} for the AdPE - two types of positional embeddings (absolute vs. relative), and two types of adversarial modes (embedding mode vs. coordinate mode). We follow the same evaluation protocol \cite{he2022masked} to report the results.

Table~\ref{tab:adv_mae+} shows the experiment results on Imagenet1K. The ViT-B with a 12-layer transformer encoder is pretrained over 400 epochs with an 8-layer transformer decoder. Then the pretrained backbone is fine-tuned end-to-end over 100 epochs with Imagenet labels, and the top-1 accuracy is reported. For comparison, the MAE and MAE+ achieve $82.95\%$ and $83.51\%$ in top-1 accuracy, respectively. 

Table~\ref{tab:ablation} reports the top-1 accuracy over 400 epochs of pretraining ViT-B under different types of positional embeddings (PE), adversarial modes, constraints, and cutoff $\epsilon$. From the ablation study in Table~\ref{tab:ablation}, we can see that the AdPE with $\ell_\infty$-constraint has higher accuracy than that with $\ell_2$-contraint for the same model type. Also, for the same type of positional embedding, the coordinate-mode adversaries perform better than the embedding-mode adversaries. We attribute this to the coordinate mode distorting image spatial structures in a more direct way with lower dimensionality (only two for x-and-y-axis) of adversaries than the embedding mode. This probably avoids the risk of over-distorting image structures arbitrarily with higher-dimensional adversaries in the embedding mode. Thus, a better balance is made to learn discriminative features from sufficiently adversarial rather than over-adversarial perturbations. The best accuracy for 400 epochs in Table~\ref{tab:ablation} is achieved by the coordinate-mode adversaries on RPE with $\epsilon = 20$ for the $\ell_\infty$-constraint, which we adopt in the following experiments. 

In Table~\ref{tab:advall}, we compare the AdPE with the other methods on Imagenet1K for pretraining ViT-B and ViT-L.  The AdPE improves the accuracy of the MAE by $0.8\%$ and $0.4\%$ on ViT-B and ViT-L without using any external datasets or models. Its accuracy also is higher than that of the MAE+ by $0.4\%$ and $0.3\%$.

We note that the AdPE is a pure MIM approach without contrastive pretraining or extra models like some other methods in Table~\ref{tab:advall}.  Particularly, contrastive pretraining often needs an additional momentum ViT branch, which makes it very slow and memory demanding in pretraining stage. For example, MoCo-v3~\cite{chen2021empirical}, a typical contrastive pretraining approach, used 128 V100 GPUs and took 10.24 GPU hours per epoch for pretraining, one order of magnitude slower than MAE and MAE+ that can be pretrained on merely eight V100 GPUs. Some approaches also resort to other models such as DALLE \cite{ramesh2021zeroshot} as an extra tokenizer model. Although the AdPE is quite flexible to further improve these approaches by adding adversarial positional embeddings, it has already outperformed them.

\subsection{Transfer Learning Results}

\begin{table}
\caption{Transfer learning results on various downstream tasks.}
\label{tab:downstream}

\centering
\begin{center}

\begin{tabular}{lcccc}
\toprule[1.5pt]

 & &   ADE20K & \multicolumn{2}{c}{COCO} \\
Measurement & Epoch& mIoU & AP$^{bbox}$ & AP$^{mask}$ \\
\midrule[1pt]
MoCo-v3~\cite{chen2021empirical}&300 & 47.3 &47.9&42.7\\
DINO~\cite{caron2021emerging}&400&47.2 &-&-\\
BEiT~\cite{bao2021beit}&800&47.1&49.8&44.4\\
CAE~\cite{chen2022context}&1600&50.2&50.0&44.0\\
iBOT~\cite{zhou2021ibot}&1600&50.0&51.2&44.2\\
SimMIM~\cite{xie2022simmim} &1600&50.0&49.1&43.8\\
MAE~\cite{he2022masked}&1600&48.1&50.3&44.9\\
\midrule
MAE+&1600&50.2&51.2&45.6\\
AdPE (ours) & 1600 &{\bf 51.5}&{\bf 53.5}&{\bf 46.5}\\

\bottomrule[1pt]
\end{tabular} 
\end{center}

\end{table}

We also conduct experiments on the transfer learning task to evaluate the generalization performance on ADE20K and COCO datasets. For a fair comparison, we still adopt the same protocol used in MAE to fine-tune the Mask R-CNN \cite{he2017mask}  end-to-end with the pretrained ViT-B backbone adapted for the FPN use on COCO \cite{lin2014microsoft}, and report AP$^{bbox}$ for object detection and AP$^{mask}$ for instance segmentation. On ADE20K, we also follow the MAE by fine-tuning UpperNet \cite{xiao2018unified} for 100 epochs with a batch size of 16. The fine-tuning learning rate is set to $0.5$ and $2e-4$ on COCO and ADE20K \cite{zhou2019semantic}, respectively.

The results in Table~\ref{tab:downstream} show that across all tasks, our model performs the best among the compared methods, which significantly improves the SOTA approaches that even adopt extra datasets and/or models. Compared to MAE~\cite{he2022masked}, MAE+ also showed better performance on those two datasets. This demonstrates its outstanding generalizability to other tasks.

\subsection{Ablation Study}
\label{sec:ablation}
\textbf{Generalization of AdPE} In Table.~\ref{tab:ablation_mae}, we further applied AdPE on 
Vanilla MAE~\cite{he2022masked} and SimMIM~\cite{xie2022simmim}. Similar improvement can also be observed. That further indicated AdPE can be a general idea for improving Masked Image Modeling (MIM) protocols.

\textbf{MAE/MAE+ with different tokens for encoder} To keep the same input information as MAE, MAE+ adopted all tokens from $112\times 112$ as encoder input. We also further analyzed different input token settings in Table.~\ref{tab:ablation_token}. As shown in Table.~\ref{tab:ablation_token}, all tokens for input can always help the improvement of finetuning performance. However, it will also increase the running time at the same time, which is more obvious for MAE with a much bigger crop. However, MAE+ can not work with masked tokens under a small crop, since 25\% of 49 tokens only include limited information, which may not be enough to train ViT.
\begin{table}[h]
\caption{Top-1 fine-tuning(FT) accuracy of AdPE and baselines over 1600 epochs of pretraining with ViT-B on Imagenet1k. }
\label{tab:ablation_token}

\begin{center}
\begin{tabular}{cccc}
  \toprule[1.5pt]
  Methods & FT & w/ AdPE & $\Delta$ \\
  \midrule[1.5pt]
 MAE~\cite{he2022masked} & 83.6 & 84.1& \better{+0.5}\\
 SimMIM~\cite{xie2022simmim} & 83.8 & 84.6 &\better{+0.8}\\
  MAE+ & 83.9 &84.4& \better{+0.5}\\
  \bottomrule[1.5pt]
\end{tabular}
\end{center}
\end{table}

\begin{table}[h]
\caption{Top-1 fine-tuning(FT) accuracy of MAE/MAE+ with different input tokens for encoder over 1600 epochs of pretraining with ViT-B on Imagenet1k. }
\label{tab:ablation_mae}

\begin{center}
\begin{tabular}{cccc}
  \toprule[1.5pt]
  Methods & Token Input & Input Size & FT\\
  \midrule[1.5pt]
 MAE~\cite{he2022masked} & Masked &$1\times 224$&83.6\\
 MAE~\cite{he2022masked} & All & $1\times 224$&83.8\\
 MAE+ & Masked & $4\times 112$& 82.8 \\
  MAE+& All & $4\times 112$& 83.9 \\
  \bottomrule[1.5pt]
\end{tabular}
\end{center}
\end{table}

\subsection{Comparison with FGSM adversaries}
\begin{table}[h]
\caption{Top-1 accuracy of FGSM over 400 epochs of pretraining with ViT-B on Imagenet1k. The FGSM is imposed on the model pretrained with the MAE+ baseline.}
\label{tab:fgsm}

\begin{center}
\begin{tabular}{ccc}
  \toprule[1.5pt]
  Constraint  $\mathcal{\ell}_{\infty}$  & FT & $\Delta_{MAE+}$ \\
  \midrule[1.5pt]
  $\epsilon=0.001$ & 83.04 &\worse{-0.47}\\
  $\epsilon=0.01$ & 83.09 &\worse{-0.42}\\
  $\epsilon=0.1$ & 82.91 &\worse{-0.60}\\
  \bottomrule[1.5pt]
\end{tabular}
\end{center}
\end{table}
The FGSM adversaries \cite{goodfellow2014explaining} are one of the most classic instance-wise attacks on deep networks.  It adds pixel-wise perturbations on the raw inputs that maximize the loss to make the worst-case attacks, and uses the resultant perturbed inputs to train the backbone network. We extend the FGSM \cite{goodfellow2014explaining} that applies instance-wise perturbations to image pixels.  It follows the classic FGSM except the MAE reconstruction loss with the multi-crop tokenization in MAE+ are adopted to compute the adversarial perturbations. Table~\ref{tab:fgsm} reports the fine-tuning results of the FGSM with various $\epsilon$ for $\ell_\infty$ constraint over $400$ epochs of pretraining ViT-B. FGSM does not perform better than MAE+ ($83.51\%$). This suggests that a straight extension of FGSM cannot improve the accuracy of a MIM-pretrained model in downstream tasks. The results also confirm the existing observation in literature \cite{tsipras2018robustness} that the instance-wise perturbations cannot improve the standard accuracy for the downstream tasks.

\subsection{Visualization of Attention Maps}
\begin{figure}[!t]
\begin{center}
\includegraphics[width=0.48\textwidth]{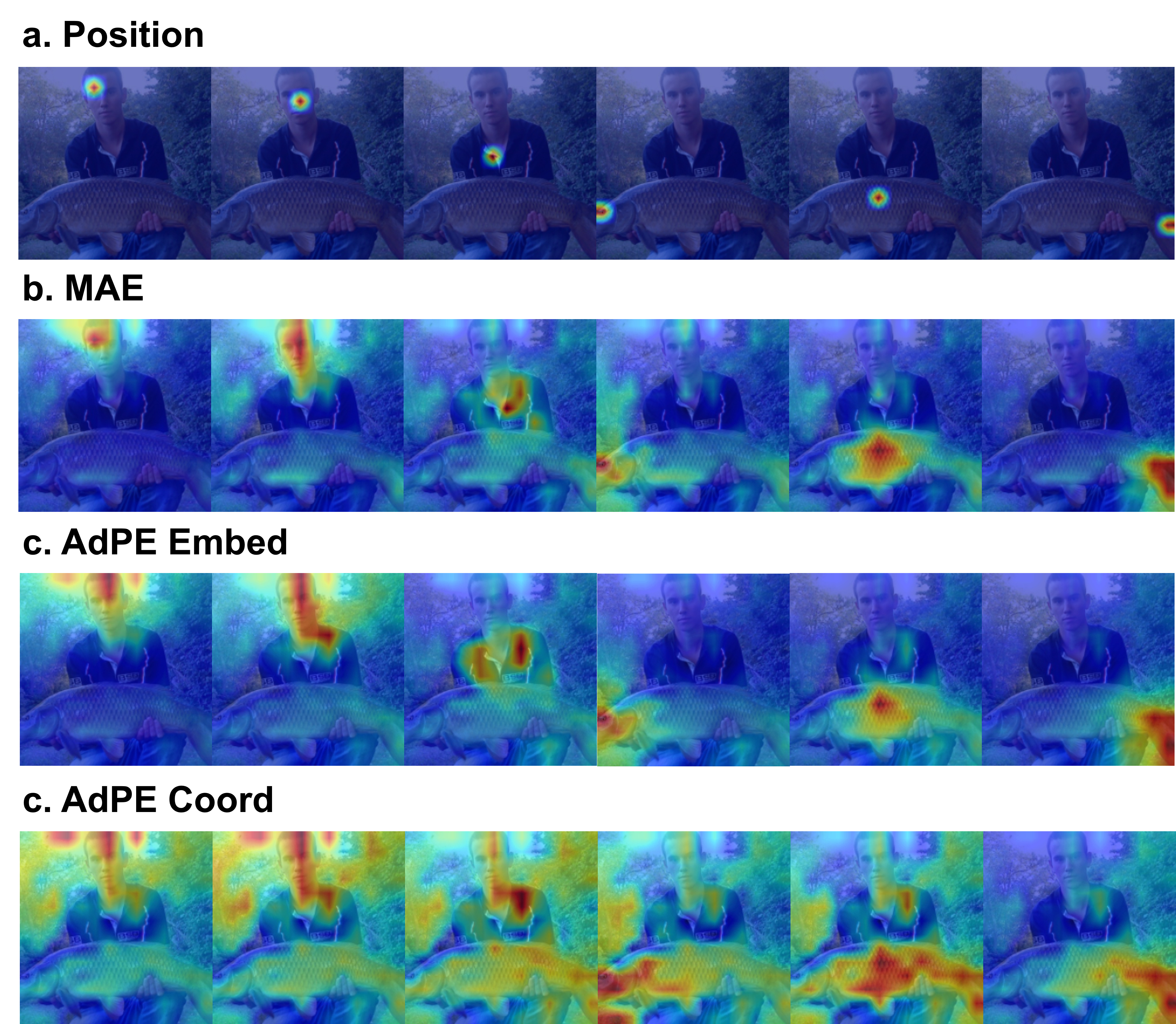}
    
\end{center}

\caption{The attention maps of the first transformer encoder layer. The MAE and AdPE are pretrained over 1600 epochs with the APE and $\epsilon=3$ for the $\ell_\infty$ constraint. }\label{fig:example_attention}

\end{figure}

\begin{figure}
\begin{center}
\includegraphics[width=0.48\textwidth]{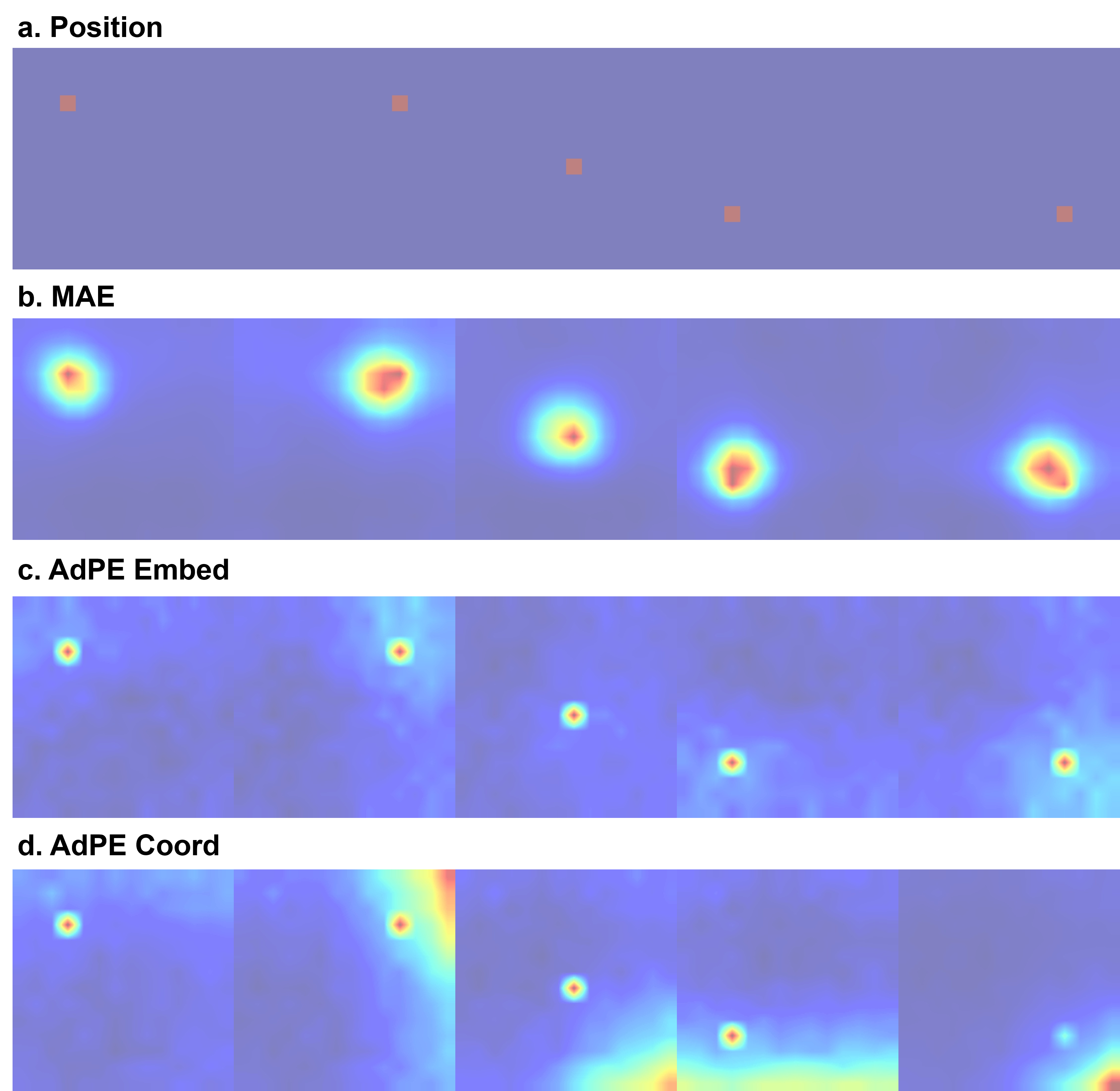}
\end{center}
\caption{The attention maps of the first transformer encoder layer averaged over $1024$ input images. Again, the MAE and AdPE are pretrained over 1600 epochs with the APE and $\epsilon=3$ for the $\ell_\infty$ constraint. }\label{fig:attention}
\end{figure}

In Fig.~\ref{fig:example_attention}, we visualize the attention maps to compare MAE and AdPE. We show that AdPE does not focus its attention over local patches to infer missing ones. Instead, it is forced to explore non-local features in a larger spatial context. Fig.~\ref{fig:attention} also visualizes the attention maps averaged over 1024 input image, with similar observations.  They show that compared to MAE, AdPE has a wider attention map covering large parts of the image. AdPE pretrained in coordinate-mode adversaries has largely distorted attention maps than that in embedding mode.  It suggests that both AdPE models cannot simply use local correlations to infer missing patches. Instead, they are forced to explore those high-level features in a larger spatial context. This verifies our assumption that the AdPE learns and integrates such high-level features from the global image context. 

\section{Conclusion}\label{sec:concl}

In this paper we present a new MAE+ baseline via multi-crop tokenization by extending the MAE-based masked image modeling.  Upon the new baseline, we show that with Adversarial Positional Embeddings (AdPE), more discriminative features can be learned from the distorted image structures by preventing a pretrained vision transformer from simply using local correlations between patches to predict masked ones. This enables the transformer to learn high-level representations generalizable to downstream tasks. We impose coordinate-mode or embedding-mode adversaries on both absolute and relative positional embeddings, and the experiment results show that the AdPE with relative positional embeddings in the coordinate mode performs the best.  We also show the AdPE has higher accuracy than the classic FGSM approach after fine-tuning the pretrained networks.

\appendices


\ifCLASSOPTIONcompsoc
  \section*{Acknowledgments}
   W. Xiao implemented the idea and performed the experiments as a remote research intern in the Seattle Research Center. Y. Wang and Z. Xuan discussed ideas and contributed to experiments. G.-J. Qi was the corresponding author (e-mail: guojunq@gmail.com), who conceived and formulated the idea as well as wrote the paper.
\else
  \section*{Acknowledgment}
\fi

\ifCLASSOPTIONcaptionsoff
  \newpage
\fi



\bibliographystyle{IEEEtran}
\bibliography{reference}

%

%




\begin{IEEEbiography}[{\includegraphics[width=1in,height=1.25in,clip,keepaspectratio]{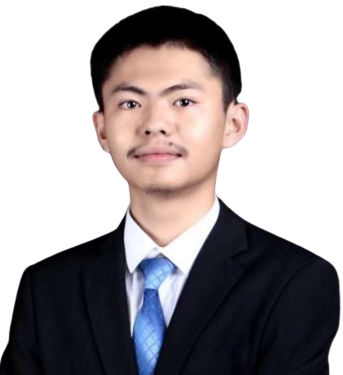}}]{Xiao Wang}
Xiao Wang received his B.S. degree from Department of Computer Science in Xi’an Jiaotong University, Xi’an, China in 2018. He is final year Ph.D. in the Department of Computer Science at Purdue University, West Lafayette, IN, USA. He is the associate editor of IEEE Transactions on Intelligent Vehicles. He has published numerous 1st-author papers in a broad range of venues: Nature Methods, Nature Communications and IEEE-TPAMI. His research interests include deep learning, computer vision, bioinformatics and intelligent systems. 
\end{IEEEbiography}

\begin{IEEEbiography}[{\includegraphics[width=1in,height=1.25in,clip,keepaspectratio]{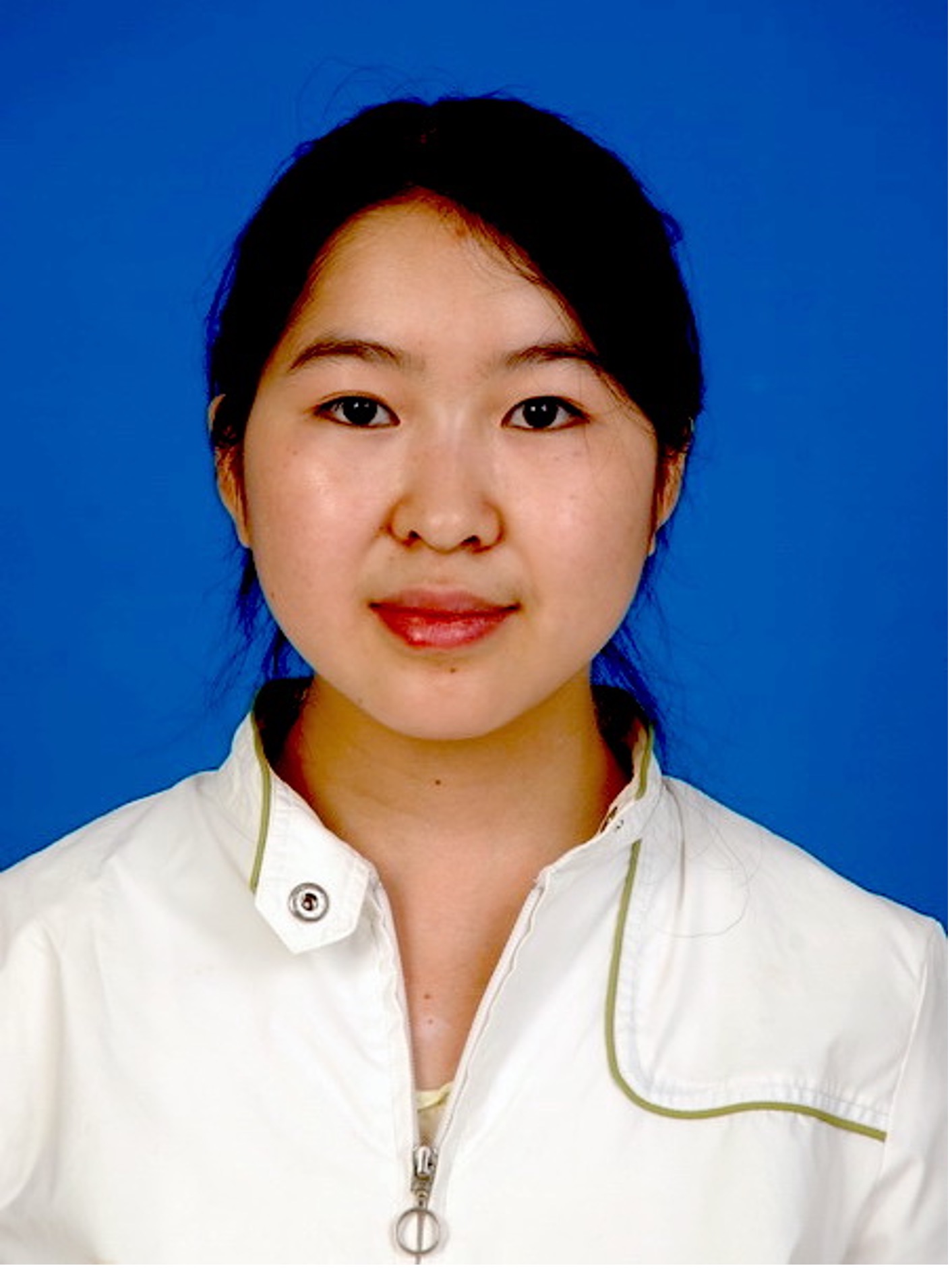}}]{Ying Wang}
Ying Wang received the PhD degree in Electrical and Computer Engineering from Texas A\&M University in 2017. Before that she was a research member in Chinese Academy of Sciences from 2010 to 2012. From Aug. 2017 to Jan. 2021 she was a senior machine learning engineer at Qualcomm AI Research Center. For the year 2021 she was a senior staff machine learning researcher at OPPO US Research Center. Since year 2022 she has been working at Amazon AWS as a senior applied scientist. Her research interests include machine learning for computer vision, autoML, semi-supervised learning and 5G wireless communications. Dr. Wang has published her works in mainstream journals and conferences including JSAC, CVPR, ECCV, NeurIPS and ISIT. Dr. Wang has over 30 patents in the field of 5G wireless communications and machine learning. Dr. Wang acted as TPC  for IEEE INFOCOM in 2018 and 2019. She has been a reviewer for a series of journals and conferences such as IEEE TCOM, IEEE Trans Circuits and Syst Video Technol. and CVPR.
\end{IEEEbiography}

\begin{IEEEbiography}[{\includegraphics[width=1in,height=1.25in,clip,keepaspectratio]{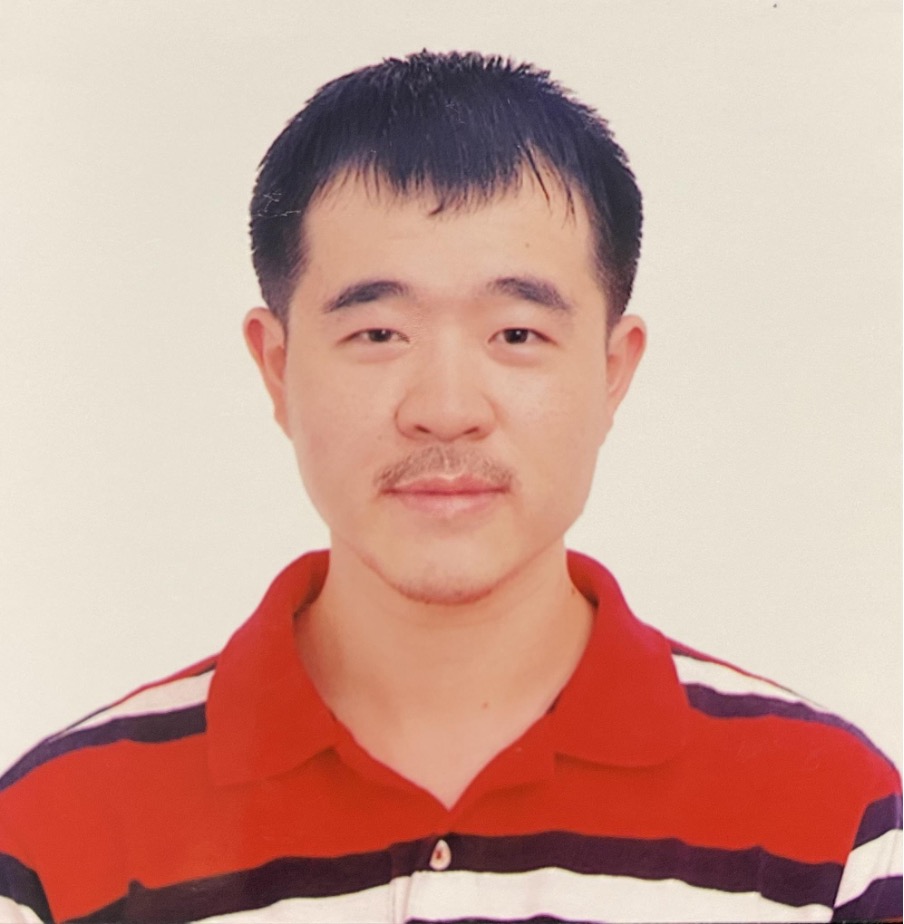}}]{Ziwei Xuan}
Ziwei Xuan received the PhD degree in Electrical and Computer Engineering from Texas A\&M University in 2022. Since the year of 2022, he has been working at OPPO US Research Center as a research scientist. His research interests include computer vision, semi-supervised learning and wireless communications.
\end{IEEEbiography}

\begin{IEEEbiography}[{\includegraphics[width=1in,height=1.25in,clip,keepaspectratio]{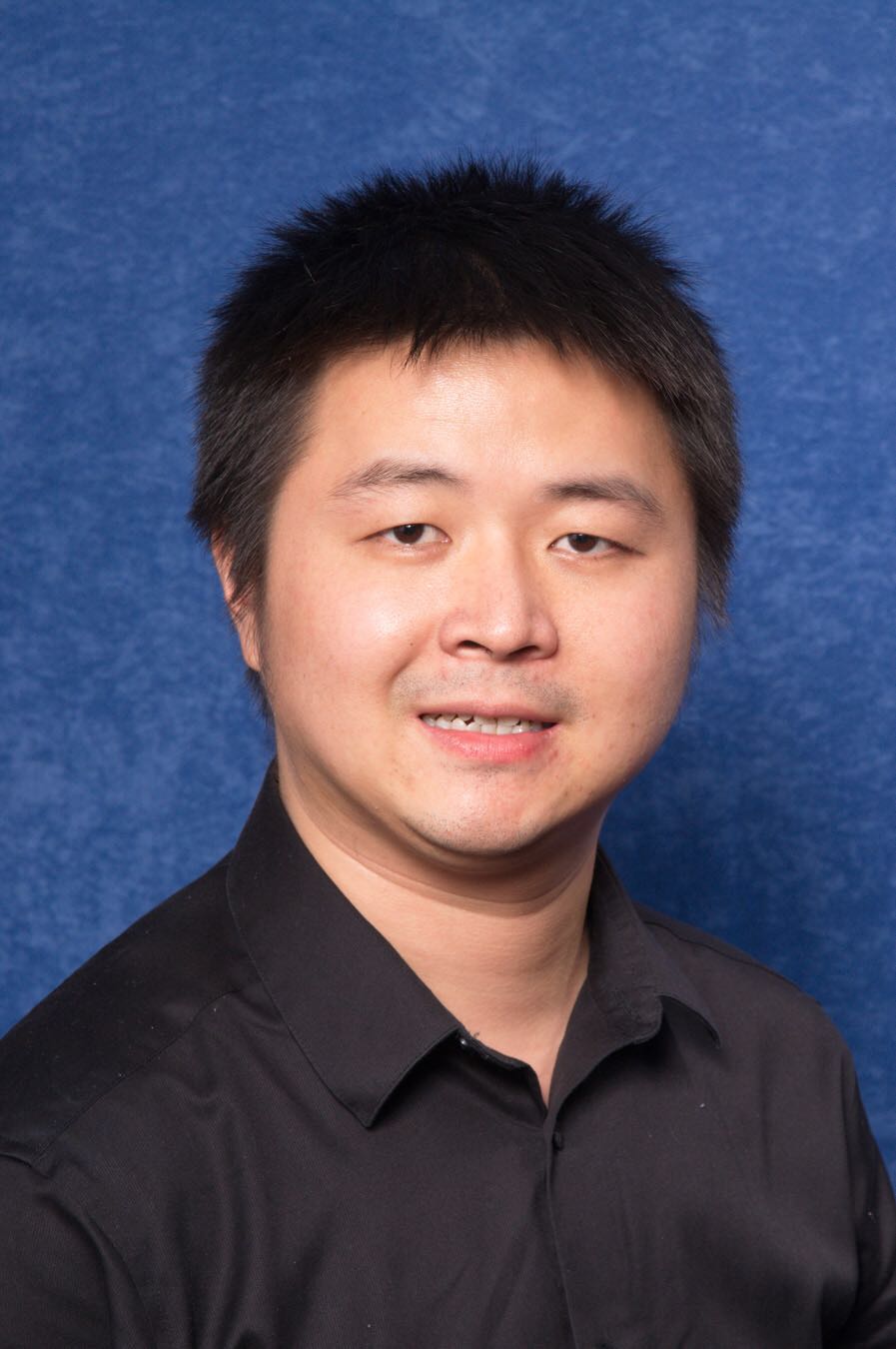}}]{Guo-Jun Qi}
Guo-Jun Qi (M14-SM18-F22) is an Adjunct Professor at Westlake University, and the Chief AI Scientist leading and overseeing an international R\&D team in the domain of Virtual Reality, Artificial Intelligence, as well as Digital Avatar Modeling and Animation at OPPO Research since 2021.
He was the Chief Scientist for multiple intelligent cloud services, including smart cities, visual computing service, medical intelligent service, and connected vehicle service at Futurewei from 2018 - 2021. He was a faculty member in the Department of Computer Science and the director of MAchine Perception and LEarning (MAPLE) Lab at the University of Central Florida since August 2014. Prior to that, he was also a Research Staff Member at IBM T.J. Watson Research Center, Yorktown Heights, NY.
Dr. Qi has published over 150 papers in a broad range of venues. Among them are the best student paper of ICDM 2014, ``the best ICDE 2013 paper" by IEEE Transactions on Knowledge and Data Engineering, as well as the best paper (finalist) of ACM Multimedia 2007 (2015). He is a Fellow of IEEE, IAPR and AAIA.
\end{IEEEbiography}

\end{document}